\documentclass[9pt,technote]{IEEEtran}
\ifCLASSINFOpdf
\else
\fi
\ifCLASSOPTIONcompsoc
 \usepackage[caption=false,font=normalsize,labelfont=sf,textfont=sf]{subfig}
\else
 \usepackage[caption=false,font=footnotesize]{subfig}
\fi

\usepackage[utf8]{inputenc} 
\usepackage[T1]{fontenc}    
\usepackage{float}
\usepackage{url}            
\usepackage{booktabs}       
\usepackage{amsfonts}       
\usepackage{nicefrac}       
\usepackage{hyperref}
\hypersetup{colorlinks}
\usepackage{graphicx}
\usepackage{amsmath,amssymb,amsopn,bm}
\usepackage{amsthm}
\usepackage{booktabs}
\usepackage{algorithm}
\usepackage{algorithmic}
\graphicspath{{./images/}}
\renewcommand{\figurename}{Fig.}
\renewcommand{\tablename}{Table}
\newcommand{\Tref}[1]{Table~\ref{#1}}
\newcommand{\Fref}[1]{Fig.~\ref{#1}}
\newcommand{\Aref}[1]{Algorithm~\ref{#1}}
\newcommand{\Sref}[1]{Section~\ref{#1}}
\newcommand{\argmin}{\mathop{\rm argmin}\limits}

\newcommand{\etal}[0]{\emph{et al}.~}

\usepackage{here}
\renewcommand{\theenumi}{\alph{enumi}}

\begin{document}
%
\title{A Novel Perspective for Positive-Unlabeled\\Learning via Noisy Labels}
%
%
%

\author{Daiki~Tanaka,
        Daiki~Ikami,
        and~Kiyoharu~Aizawa,
\IEEEcompsocitemizethanks{\IEEEcompsocthanksitem D. Tanaka is with the University of Tokyo.
e-mail: tanaka@hal.t.u-tokyo.ac.jp
\IEEEcompsocthanksitem D. Ikami is with NTT Communication Science Laboratories.
\IEEEcompsocthanksitem K. Aizawa is with the University of Tokyo.}
}

\maketitle

\begin{abstract}
Positive-unlabeled learning refers to the process of training a binary classifier using only positive and unlabeled data. Although unlabeled data can contain positive data, all unlabeled data are regarded as negative data in existing positive-unlabeled learning methods, which resulting in diminishing performance. We provide a new perspective on this problem -- considering unlabeled data as noisy-labeled data, and introducing a new formulation of PU learning as a problem of joint optimization of noisy-labeled data. This research presents a methodology that assigns initial pseudo-labels to unlabeled data which is used as noisy-labeled data, and trains a deep neural network using the noisy-labeled data. Experimental results demonstrate that the proposed method significantly outperforms the state-of-the-art methods on several benchmark datasets.
\end{abstract}


%
\IEEEpeerreviewmaketitle

\section{Introduction}
Positive-unlabeled (PU) learning refers to the process of training a binary classifier by utilizing only positive (P) and unlabeled (U) data. In such a problem setting, the unlabeled data class prior probability is known.
Research on PU learning was initiated by~\cite{denis1998pac,de1999positive,letouzey2000learning},
and it is expected to be mainly applied to retrieval and outlier detection~\cite{elkan2008learning,ward2009presence,scott2009novelty,blanchard2010semi}.

Existing PU learning can be roughly divided into two types depending on the use of unlabeled data: reweighting approaches and two-step approaches.

Reweighting approaches regard all unlabeled data as negative samples and train a classifier by weighted training samples.
For example, biased SVM~\cite{liu2003building} and weighted logistic regression~\cite{lee2003learning} incur different costs for misclassified positive and negative samples.

The representative two-step approaches first select a set of reliable negative samples from unlabeled data and then train a classifier by applying a traditional supervised positive-negative (PN) learning. However, previous works~\cite{liu2002partially,li2003learning} achieve inferior performance as compared to recent single-step approaches because of the incorrect identification of negative samples.

Both approaches require the weight parameters to be manually adjusted for labeled and unlabeled data, which is computationally expensive.

The ability to circumvent the need for manual tuning of the parameters increased the popularity of unbiased PU learning and its extensions~\cite{du2014analysis,du2015convex}, and achieving state-of-the-art performance in PU learning by the use of deep neural networks~\cite{kiryo2017positive,hsieh2018classification}.
Although reweighting approaches have shown success in PU learning, performance degradation may have been caused by treating all unlabeled data as negative, because unlabeled data contains both positive and negative samples.

In this paper, we shed light upon on PU learning -- we have treated unlabeled data as noisy negative data and formulated PU learning by a joint optimization of learning DNN parameters and unlabeled data.
We have been inspired by one of the most effective methods,
the joint optimization framework~\cite{tanaka2018joint}, in which a deep neural network and noisy labels are alternately optimized.

Simply applying this framework to PU learning does not work well because clean-labeled data consists of only positive samples. We have introduced a new weighting parameter $\lambda$, which makes the network focus on clean data as compared to noisy data in the beginning of training.

We have also introduced a new way to assign initial pseudo-labels -- it is not known if considering all unlabeled data as negative is the best strategy for initial assigned label assignment.
Let $1$ denote positive labels and $0$ denote negative ones. If we choose not to restrict to only a hard-labeling, $\{0,1\}$, but also allow for soft-labels, $[0,1]$, we suppose that initially assigned labels should not be set to a negative hard-label, but instead, should be equal to the class prior probability, in order to minimize the Kullback-Leibler divergence from initial labels to the ground-truth labels. We have experimentally found that such an assignment prevents a deep neural network from overfitting and that the formulation works well.
The main contributions of this study are as follows.
\begin{itemize}
    \item We have introduced a novel approach; the problem setting of PU learning can be regarded as that of learning with noisy-labeled data.
    We have formulated a new method for PU learning such that the network parameters and noisy labels are jointly optimized.
    \item Unlike most existing approaches that assign negative labels to all the unlabeled data, we determine initial labels based on the class prior probability. We have experimentally confirmed that our initial label assignment achieved better performance compared to the all-negative label approach if applied to the proposed method.
    \item We have evaluated the proposed method on several benchmark datasets. Experimental results demonstrates that our method significantly outperforms previous state-of-the-art PU learning approaches.
\end{itemize}

\section{Related Works}
In this section, we first state the problem settings of PU learning~\cite{kiryo2017positive}, and then we introduce the main two types of PU learning, namely reweighting approaches and two-step approaches. Finally, we explain the details about learning with noisy-labeled data because we consider unlabeled data assigned pseudo-labels as noisy-labeled data.
\subsection{Problem settings of PU learning}
Let $X\in\mathbb{R}^d$ and $Y\in\{0,1\}$ be the input and output random variables, respectively. Let $p(x,y)$ be the joint density of $(X,Y)$, $p_\mathrm{p}(x)=p(x|Y=1)$ and $p_\mathrm{n}(x)=p(x|Y=0)$ be the P and N marginals, $p(x)$ be the U marginal, $\pi_\mathrm{p}=p(Y=1)$ be the class prior probability, and $\pi_\mathrm{n}=p(Y=0)=1-\pi_\mathrm{p}$.
We regard $\pi_\mathrm{p}$ as known throughout this paper, although some studies estimate it from P and U data~\cite{menon2015learning,ramaswamy2016mixture,jain2016estimating,christoffel2016class}.
P and U data are sampled independently from $p_\mathrm{p}(x)$ and $p(x)$ as $\mathcal{X}_\mathrm{p}=\{x^\mathrm{p}_i\}^{n_\mathrm{p}}_{i=1}\sim p_\mathrm{p}(x)$ and
$\mathcal{X}_\mathrm{u}=\{x^\mathrm{u}_i\}^{n_\mathrm{u}}_{i=1}\sim p(x)$, respectively.
PU learning means training a classifier from $\mathcal{X}_\mathrm{p}$ and $\mathcal{X}_\mathrm{u}$ unlike PN learning in which a classifier is usually trained from $\mathcal{X}_\mathrm{p}$ and $\mathcal{X}_\mathrm{n}=\{x^\mathrm{n}_i\}^{n_\mathrm{n}}_{i=1}\sim p_\mathrm{n}(x)$.

\subsection{Reweighting approaches}
In this subsection, we describe two recent approaches of reweighting approaches: unbiased PU learning~\cite{du2014analysis,du2015convex} and non-negative PU learning~\cite{kiryo2017positive}.

\subsubsection{Unbiased PU learning}
Let $\sigma:\mathbb{R}^d\rightarrow[0,1]$ denote the network in which the final layer is the sigmoid function. The loss to be minimized is $L(\sigma)=\mathbb{E}_{(X,Y)\sim p(x,y)}[Y\sigma(X)+(1-Y)(1-\sigma(X))]=\pi_\mathrm{p}L^1_\mathrm{p}(\sigma)+\pi_nL^0_\mathrm{n}(1-\sigma)$,
where $L^1_\mathrm{p}(\sigma)=\mathbb{E}_{X\sim p_\mathrm{p}}[\sigma(X)]$ and
$L^0_\mathrm{n}(\sigma)=\mathbb{E}_{X\sim p_\mathrm{n}}[1-\sigma(X)]$.
In PN learning, $L(\sigma)$ can be directly approximated as follows:
\begin{equation}\label{eq:pn}
    \hat{L}_{\mathrm{pn}}(\sigma)=\pi_\mathrm{p}\hat{L}^1_\mathrm{p}(\sigma)+\pi_\mathrm{n}\hat{L}^0_\mathrm{n}(\sigma),
\end{equation}
where $\hat{L}^1_\mathrm{p}(\sigma)=\frac{1}{n_\mathrm{p}}\sum^{n_\mathrm{p}}_{i=1}\sigma(x^\mathrm{p}_i)$,
$\hat{L}^0_\mathrm{n}(\sigma)=\frac{1}{n_\mathrm{n}}\sum^{n_\mathrm{n}}_{i=1}(1-\sigma(x^\mathrm{n}_i))$.
In PU learning, $\mathcal{X}_\mathrm{n}$ cannot be used and thus, unbiased PU learning~\cite{du2014analysis,du2015convex} indirectly approximates $L(\sigma)$. As $\pi_\mathrm{n}p_\mathrm{n}(x)=p(x)-\pi_\mathrm{p}p_\mathrm{p}(x)$,
$\pi_\mathrm{n}L^0_\mathrm{n}(\sigma)=L^0_\mathrm{u}(\sigma)-\pi_\mathrm{p}L^0_\mathrm{p}(\sigma)$ is formulated, where
$L^0_\mathrm{p}(\sigma)=\mathbb{E}_{X\sim p_\mathrm{p}}[1-\sigma(X)]$ and
$L^0_\mathrm{u}(\sigma)=\mathbb{E}_{X\sim p(x)}[1-\sigma(X)]$.
Therefore, $L(\sigma)$ can be indirectly approximated as follows:
\begin{equation}\label{eq:upu}
    \hat{L}_{\mathrm{pu}}(\sigma)=\pi_\mathrm{p}\hat{L}^1_\mathrm{p}(\sigma)+\hat{L}^0_\mathrm{u}(\sigma)-\pi_\mathrm{p}\hat{L}^0_\mathrm{p}(\sigma),
\end{equation}
where $\hat{L}^0_\mathrm{u}(\sigma)=\frac{1}{n_\mathrm{u}}\sum^{n_\mathrm{u}}_{i=1}(1-\sigma(x^\mathrm{u}_i))$ and
$\hat{L}^0_\mathrm{p}(\sigma)=\frac{1}{n_\mathrm{p}}\sum^{n_\mathrm{p}}_{i=1}(1-\sigma(x^\mathrm{p}_i))$.

\subsubsection{Non-negative PU learning}
By using unbiased PU learning, we can obtain $\hat{g}_{\mathrm{pu}}$ which minimizes $\hat{L}_{\mathrm{pu}}(g)$; however, it is experimentally found that the loss will be negative if the model is significantly flexible, such as a deep neural network~\cite{kiryo2017positive}. This problem is solved by non-negative PU learning~\cite{kiryo2017positive}, which modifies the empirical loss as follows:
\begin{equation}\label{eq:nnpu}
    \tilde{L}_{\mathrm{pu}}(\sigma)=\pi_\mathrm{p}\hat{L}^1_\mathrm{p}(\sigma)+\max\left\{0,\hat{L}^0_\mathrm{u}(\sigma)-\pi_\mathrm{p}\hat{L}^0_\mathrm{p}(\sigma)\right\}.
\end{equation}
Gradient ascent is performed along $\nabla(\hat{L}^0_\mathrm{u}(\sigma)-\pi_\mathrm{p}\hat{L}^0_\mathrm{p}(\sigma))$ when $\hat{L}^0_\mathrm{u}(\sigma)-\pi_\mathrm{p}\hat{L}^0_\mathrm{p}(\sigma)$ becomes smaller than some threshold value.
This modification prevents the model from overfitting.

\subsection{Two-step approaches}
The two-step approach consists of the following two steps: i) assigning pseudo-labels to unlabeled data and ii) learning from labeled and pseudo-labeled data. The classical approaches only identify reliable negative samples~\cite{liu2003building,lee2003learning} for the first step and do not perform well because of the inaccurate identification of negative samples. Recently, Hsieh \etal have applied non-negative PU learning, which is the best performing reweighting approach, as the first step, and achieves state-of-the-art performance~\cite{hsieh2018classification}. Our method can also be categorized into this group: we regard unlabeled data as noisy negative data and perform a joint optimization for learning from noisy labels to identify both negative and positive samples from unlabeled data.

\subsection{Learning with noisy-labeled data}
The studies on noisy-labeled data can be divided into two major categories. The first category studies utilize the modification of the cross-entropy loss to prevent a classifier from overfitting to noisy-labeled data. Some studies use a noise-transition matrix~\cite{patrini2016making,sukhbaatar2014training,jindal2016learning,vahdat2017toward},
and others use the mean-absolute-error~\cite{ghosh2017robust,zhang2018generalized}.

The studies from the second category are of the Bootstrapping type~\cite{reed2014training}, in which labels are iteratively updated by the predictions of the model during training. The joint optimization framework~\cite{tanaka2018joint} uses this scheme, and demonstrates better performance because a high learning rate prevents a classifier from overfitting to noisy-labeled data. This framework is one of the most effective approaches for high noisy-labeled data; it works well even if 90 \% of the labels are randomized. The research introduced above inspired the authors to solve PU learning based on a joint optimization approach.

\section{Method}
\begin{figure}[t]
    \centering
    \includegraphics[width=9cm]{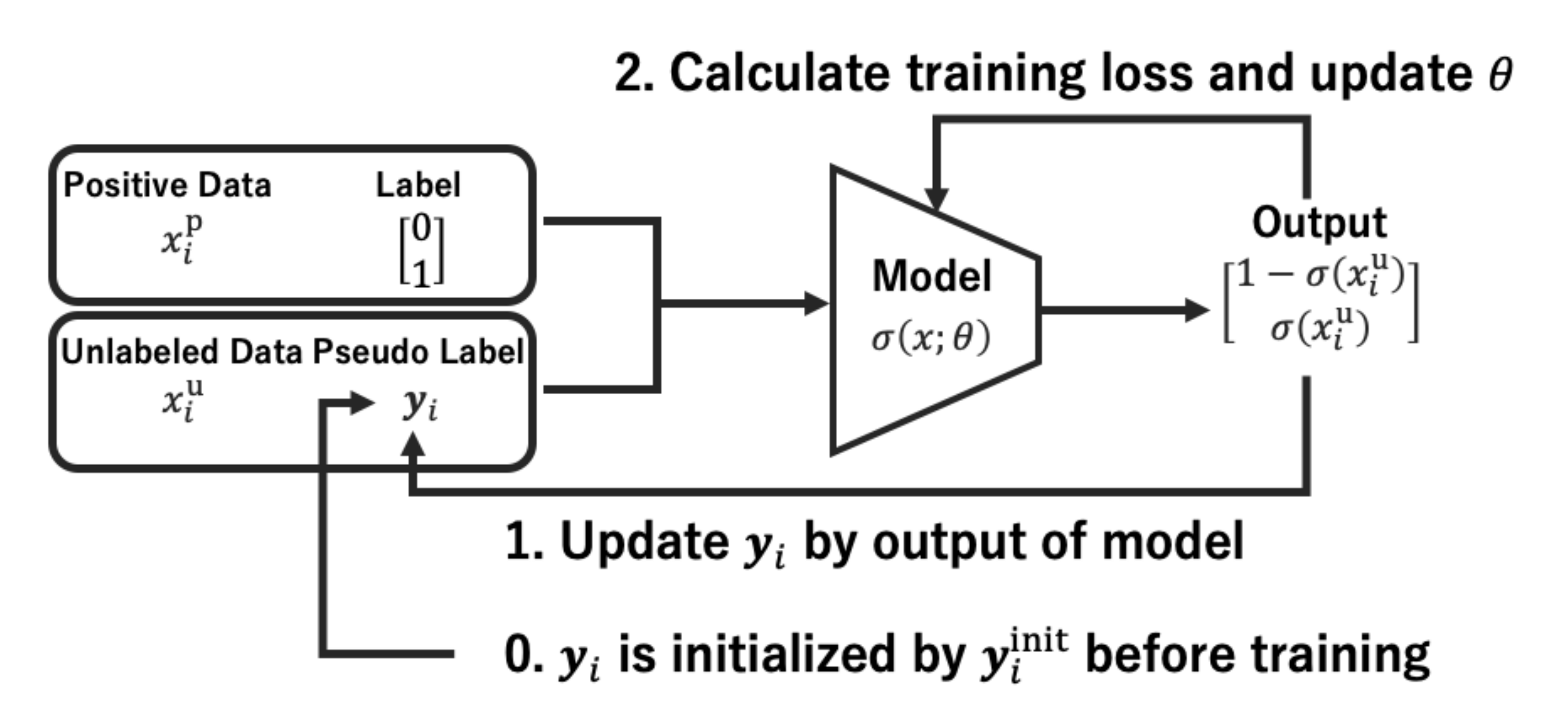}
    \caption{The concept of our proposed method. Initialized pseudo labels are assigned to unlabeled data, and updated by outputs of the model for every epoch.}
    \label{fig:method}
\end{figure}
\begin{algorithm*}[t]
    \caption{Joint optimization for PU learning}
    \label{alg1}
    \begin{algorithmic}[1]
        \REQUIRE{training data $(\mathcal{X}_\mathrm{p},\mathcal{X}_\mathrm{u},\mathcal{Y})$}
        \REQUIRE{hyperparameters $\lambda_{\mathrm{init}},r,e_{\mathrm{start}},e_{\mathrm{end}}$}
    \REQUIRE{matrix $\mathcal{Z}(n_\mathrm{u}\times e_{\mathrm{end}})$}
        \ENSURE{model parameter $\theta$ for $\hat{\sigma}_{\mathrm{joint}}(x;\theta)$ minimizing the validation loss}
        \STATE Let $\mathcal{A}$ be an external SGD-like stochastic optimization algorithm such as AMSGrad~\cite{reddi2018convergence}
        \STATE $\mathcal{Y}\leftarrow\mathcal{Y}^{\mathrm{init}}$ (Initialize labels of unlabeled data by pseudo-labels, each of which is equal to the class prior probability $\pi_\mathrm{p}$)
        \FOR{$i=1$ \textbf{to} $e_{\mathrm{end}}$}
    \STATE $\lambda\leftarrow \frac{e_\mathrm{end}-i}{e_\mathrm{end}-1}(\lambda_{\mathrm{init}}-\frac{n_\mathrm{p}}{n_\mathrm{u}})+\frac{n_\mathrm{p}}{n_\mathrm{u}}$
        \STATE Shuffle $(\mathcal{X}_\mathrm{p},\mathcal{X}_\mathrm{u},\mathcal{Y},\mathcal{Z})$ into $N$ mini-batches, and denote by $(\mathcal{X}^j_\mathrm{p},\mathcal{X}^j_\mathrm{u},\mathcal{Y}^j,\mathcal{Z}^j)$ the $j$-th mini-batch
        \FOR{$j=1$ \textbf{to} $N$}
        \STATE Set gradient $\nabla_\theta \hat{L}_{\mathrm{joint}}(\sigma;\mathcal{X}^j_\mathrm{p},\mathcal{X}^j_\mathrm{u},\mathcal{Y}^j)$
    \FOR{$k=1$ \textbf{to} batchsize}
    \STATE $z^{jk}_i\leftarrow \sigma(x^{jk}_\mathrm{u})$ (Preserve predictions of the model)
        \IF{$i\geq e_{\mathrm{start}}$}
        \STATE $y^{jk}\leftarrow \frac{1}{r}\sum^i_{l=i-r+1}z^{jk}_l$ (Update labels of unlabeled data)
        \ENDIF
    \ENDFOR
        \STATE Update $\theta$ by $\mathcal{A}$ with its current step size $\eta$
        \ENDFOR
        \ENDFOR
    \end{algorithmic}
\end{algorithm*}
In this section, we introduce the joint optimization for learning from positive-unlabeled data. The method introduced first assigns pseudo-labels to unlabeled data, and then iteratively updates pseudo-labels, as shown in \Fref{fig:method}. We first explain how to update pseudo-labels; then we will describe the process of assigning initial pseudo-labels in \Sref{sec:init}.
\subsection{Joint Optimization}
Let $\mathcal{Y}=\{y_i\in[0,1]\}^{n_\mathrm{u}}_{i=1}$ be pseudo-labels assigned to $\mathcal{X}_\mathrm{u}$, which may contain incorrect labels. The goal is to train a network from clean positive data $\mathcal{X}_\mathrm{p}$ and  pseudo-labeled data $\mathcal{X}_\mathrm{u}$. 


In the joint optimization, pseudo-labels $\mathcal{Y}$ and the network $\sigma$ are alternately updated, as shown in~\Aref{alg1}.
As a result of the joint optimization, the noisy pseudo-labeled data is updated to more precise labels. 
Matrix $\mathcal{Z}$ is required to preserve the predictions of the model, and the labels are updated by the average predictions in the last $r$ epochs. The total empirical loss $\hat{L}_{\mathrm{joint}}$ is constructed by three terms as follows:
\begin{equation}\label{eq:total}
    \hat{L}_{\mathrm{joint}}(\sigma)=\hat{L}_{\mathrm{class}}(\sigma)+\alpha\hat{L}_{\mathrm{reg1}}(\sigma)+\beta\hat{L}_{\mathrm{reg2}}(\sigma),
\end{equation}
where $\hat{L}_{\mathrm{class}}(\sigma)$, $\hat{L}_{\mathrm{reg1}}(\sigma)$, and $\hat{L}_{\mathrm{reg2}}(\sigma)$ denote the classification loss and two regularization losses, respectively, and $\alpha$ and $\beta$ denote hyperparameters.

Unlike the study regarding noisy-labeled data by~\cite{tanaka2018joint}, we deal with binary-classification with PU data and thus, we formulate $\hat{L}_{\mathrm{class}}(\sigma)$ as follows:
\begin{equation}
\resizebox{.91\linewidth}{!}{$
\displaystyle
\begin{split}
    &\ \ \ \  \hat{L}_{\mathrm{class}}(\sigma)=\lambda\hat{L}^1_\mathrm{p}(\sigma)+\hat{L}_\mathrm{noisy}(\sigma)\\
    &=\lambda\hat{L}^1_\mathrm{p}(\sigma)+\frac{1}{n_\mathrm{u}}\sum^{n_\mathrm{u}}_{i=1}D_{\mathrm{KL}}\left(
    \begin{array}{c||c}
    \left[
    \begin{array}{c}
    y_i \\
    1-y_i
    \end{array}
    \right]
    &
    \left[
    \begin{array}{c}
    \sigma(x^\mathrm{u}_i) \\
    1-\sigma(x^\mathrm{u}_i)
    \end{array}
    \right]
    \end{array}
    \right),
\end{split}
$}
\end{equation}
where we manipulate $\lambda$ to gradually decrease during training because positive clean data is more beneficial for training than unlabeled data in the early phase of training.
In this study, $\lambda$ is reduced linearly, and it will finally be equal to $\frac{1}{n_\mathrm{u}}$. Note that the joint optimization framework does not work well if $\lambda$ is set to the constant value which is equal to the class prior probability like other PU learning studies.

Following~\cite{tanaka2018joint}, two regularization losses are defined as follows:
\begin{equation}
\resizebox{.91\linewidth}{!}{$
\displaystyle
    \hat{L}_{\mathrm{reg1}}(\sigma)=D_{\mathrm{KL}}\left(
    \begin{array}{c||c}
    \left[
    \begin{array}{c}
    \pi_\mathrm{p} \\
    \pi_\mathrm{n}
    \end{array}
    \right]
    &
    \left[
    \begin{array}{c}
    \frac{1}{n_\mathrm{u}}\sum^{n_\mathrm{u}}_{i=1}\sigma(x^\mathrm{u}_i) \\
    \frac{1}{n_\mathrm{u}}\sum^{n_\mathrm{u}}_{i=1}(1-\sigma(x^\mathrm{u}_i))
    \end{array}
    \right]
    \end{array}
    \right),
    $}
\end{equation}
\begin{equation}
\begin{split}
    \hat{L}_{\mathrm{reg2}}(\sigma)&=\frac{1}{n_\mathrm{u}}\sum^{n_\mathrm{u}}_{i=1}\bigl(\sigma(x^\mathrm{u}_i)\log\sigma(x^\mathrm{u}_i) \\
    &+(1-\sigma(x^\mathrm{u}_i))\log(1-\sigma(x^\mathrm{u}_i))\bigr).
\end{split}
\end{equation}
Utilizing previously defined loss terms, we can avert the trapping into undesirable pseudo-labels.
The first regularization loss term $\hat{L}_{\mathrm{reg1}}(\sigma)$ requires the ground-truth class prior probability, and we consider that it is available in the problem setting of PU learning.

\subsection{Assigning initial pseudo-labels}\label{sec:init}
The method introduced requires initial pseudo-labels for unlabeled data. A straightforward solution is assigning negative labels to all the unlabeled data in a similar way to the existing methods. However, this assignment ignores the class prior probability $\pi_\mathrm{p}$.
Let we consider the ground-truth labels of unlabeled data, $\mathcal{Y}^{\mathrm{GT}}=\{y^{\mathrm{GT}}_i\in\{0,1\}\}^{n_\mathrm{u}}_{i=1}$.
From the definition of $\pi_\mathrm{p}$, the ground-truth labels satisfy $\frac{1}{n_\mathrm{u}}\sum^{n_\mathrm{u}}_{i=1}y^{\mathrm{GT}}_i = \pi_\mathrm{p}$. Further, we assume that all the initial labels are the same because we cannot distinguish the unlabeled data. Then, we obtain the initial label $y_{\mathrm{init}}$ by minimizing the KL divergence from $\mathcal{Y}$ to $\mathcal{Y}^{\mathrm{GT}}$:
\begin{equation}
\resizebox{.91\linewidth}{!}{$
\displaystyle
    \begin{split}
    y_{\mathrm{init}} &= \argmin_{y}\frac{1}{n_\mathrm{u}}\sum^{n_\mathrm{u}}_{i=1}D_{\mathrm{KL}}\left(
    \begin{array}{c||c}
    \left[
    \begin{array}{c}
    y^{\mathrm{GT}}\\
    1-y^{\mathrm{GT}}
    \end{array}
    \right]
    &
    \left[
    \begin{array}{c}
    y\\
    1-y
\end{array}
\right]
    \end{array}
    \right)
    \\
    &=\argmin_y\left(-\frac{1}{n_\mathrm{u}}\sum^{n_\mathrm{u}}_{i=1}\bigl(y^{\mathrm{GT}}\log y+(1-y^{\mathrm{GT}})\log(1-y)\bigr)\right)\\
&=\argmin_y\bigl(-\pi_\mathrm{p}y-\pi_\mathrm{n}(1-y)\bigr)\\
&=\pi_\mathrm{p},
\end{split}
$}
\end{equation}
where $y^{\mathrm{GT}}_i\log y^{\mathrm{GT}}_i=(1-y^{\mathrm{GT}}_i)\log(1-y^{\mathrm{GT}}_i)=0$.
Unless otherwise specified, we use $y_i = \pi_\mathrm{p}$ for all $i$ as the initial labels.

\section{Experiments}\label{sec:experiment}
We have verified the effectiveness of the proposed method using three benchmark datasets: MNIST, CIFAR-10, and 20 Newsgroups, which are commonly used for evaluation of PU learning in~\cite{kiryo2017positive,hsieh2018classification}.

\subsection{Datasets}
\textbf{MNIST}: MNIST~\cite{lecun1998gradient} is a gray-scale image dataset. We set 0, 2, 4, 6, and 8 as the positive class and 1, 3, 5, 7, and 9 as the negative class, such that the class prior probability is $\pi_\mathrm{p}=0.49$.

\noindent\textbf{CIFAR-10}: CIFAR-10~\cite{krizhevsky2009learning} is an RGB image dataset, and we define two patterns of the positive class according to~\cite{hsieh2018classification}. The first set includes airplane, automobile, ship, and truck as the positive class and bird, cat, deer, dog, frog, and horse as the negative class in order to distinguish the vehicles from the animals, such that the class prior probability is $\pi_\mathrm{p}=0.4$. The second set includes cat, deer, dog, and horse as the positive class and airplane, automobile, ship, truck, bird, and frog as the negative class in order to distinguish the mammals from the non-mammals, such that the class prior probability is $\pi_\mathrm{p}=0.4$.

\noindent\textbf{20 Newsgroups}: 20 Newsgroups~\cite{lang1995newsweeder} is a text dataset. We set alt., comp., misc., and rec. as the positive class and sci., soc., and talk. as the negative class, such that the class prior probability is $\pi_\mathrm{p} = 0.56$.

\subsection{Implementation details}
We used the standard test examples, such that the test set size is 10000 for MNIST and CIFAR-10, and 7528 for 20 Newsgroups. With respect to the training set, we sample 500 as positive and 6000 as unlabeled for MNIST and 20 Newsgroups, respectively, and 1000 as positive and 10000 as unlabeled for CIFAR-10. The validation set is always one-fifth of the training set.

The validation set is used for both tuning the hyperparameters and choosing the model parameters with the lowest validation loss among those calculated by \eqref{eq:nnpu} after every epoch.

For MNIST, we used a four-layer convolutional neural network and the hyperparameters were set as $\mathrm{lr}=0.005$, $r=10$, $e_{\mathrm{start}}=20$, $\lambda_{\mathrm{init}}=10.0$, $\alpha=10.0$, $\beta=2.0$, and $e_{\mathrm{end}}=100$. We changed some hyperparameters as $\lambda_{\mathrm{init}}=10.0$, $\alpha=10.0$ for $\pi_\mathrm{p}\in\{0.6, 0.7\}$.

For CIFAR-10, we used a PreAct ResNet-18~\cite{he2016identity}, and the hyperparameters were set as $\mathrm{lr}=0.001$, $r=10$, $e_{\mathrm{start}}=20$, $\lambda_{\mathrm{init}}=0.5$, $\alpha=2.0$, $\beta=0.5$, and $e_{\mathrm{end}}=200$.

For 20 Newsgroups, we preprocessed the raw text data into 9216-dimensional feature vectors by the pre-trained ELMoword embedding~\cite{peters2018deep} from AllenNLP~\cite{Gardner2017AllenNLP}\footnote{See \url{https://allennlp.org/elmo}}, as suggested by~\cite{ruckle2018concatenated}.
We used a three-layer fully connected neural network, and the hyperparameters were set as $\mathrm{lr}=0.01$, $r=10$, $e_{\mathrm{start}}=10$, $\lambda_{\mathrm{init}}=2.0$, $\alpha=2.0$, $\beta=2.0$, and $e_{\mathrm{end}}=50$.

\begin{table*}[t]
    \caption{Mean and standard deviation of test error rates over 10 trials for MNIST, CIFAR-10, and 20 Newsgroups. Different assignments of the initial pseudo-labels for the unlabeled data are compared using the same 10 random samplings. The best results are highlighted in bold.}
    \begin{center}
    \begin{tabular}{c|c|c|c}\hline
    \multicolumn{1}{c|}{Dataset}&\multicolumn{1}{c|}{All negative}&\multicolumn{1}{c|}{Randomized hard-labels}&\multicolumn{1}{c}{Proposed assignment}\\\hline
    \multicolumn{1}{c|}{MNIST}&$4.65\pm0.35$&$6.18\pm0.83$&$\bm{3.69}\pm\bm{0.69}$\\\hline
    \multicolumn{1}{c|}{\begin{tabular}{c}CIFAR-10\vspace{-0.5mm}\\(vehicles as P)\end{tabular}}&$10.16\pm0.32$&$10.15\pm0.43$&$\bm{9.84}\pm\bm{0.30}$\\\hline
    \multicolumn{1}{c|}{\begin{tabular}{c}CIFAR-10\vspace{-0.5mm}\\(mammals as P)\end{tabular}}&$21.03\pm0.76$&$20.92\pm1.23$&$\bm{20.21}\pm\bm{0.70}$\\\hline
    \multicolumn{1}{c|}{20 Newsgroups}&$15.94\pm1.76$&$26.60\pm3.21$&$\bm{13.09}\pm\bm{0.72}$\\\hline
    \end{tabular}
    \end{center}
    \label{tab:results0}
\end{table*}

\begin{table}[t]
    \caption{Mean and standard deviation of test error rates over 10 trials for MNIST, CIFAR-10, and 20 Newsgroups. Different methods are compared using the same 10 random samplings. }
    \begin{center}
    \begin{tabular}{c|c|c|c}\hline
    \multicolumn{1}{c|}{Dataset}&\multicolumn{1}{c|}{nnPU}&\multicolumn{1}{c|}{PUbN\textbackslash N}&\multicolumn{1}{c}{Our method}\\\hline
    \multicolumn{1}{c|}{MNIST}&$6.64\pm1.33$&$4.62\pm0.58 $&$\bm{3.69}\pm\bm{0.69}$\\\hline
    \multicolumn{1}{c|}{\begin{tabular}{c}CIFAR-10\vspace{-0.5mm}\\(vehicles as P)\end{tabular}}&$12.21\pm0.68$&$10.86\pm0.31$&$\bm{9.84}\pm\bm{0.30}$\\\hline
    \multicolumn{1}{c|}{\begin{tabular}{c}CIFAR-10\vspace{-0.5mm}\\(mammals as P)\end{tabular}}&$22.57\pm1.01$&$21.26\pm0.39$&$\bm{20.21}\pm\bm{0.70}$\\\hline
    \multicolumn{1}{c|}{20 Newsgroups}&$15.00\pm0.72$&$13.98\pm0.78$&$\bm{13.09}\pm\bm{0.72}$\\\hline
    \end{tabular}
    \end{center}
    \label{tab:results}
\end{table}

\subsection{Results}
\subsubsection{Evaluation of different initial pseudo-labels}
In the first experiment, we trained the neural network models from different initial pseudo-labels. We compared the proposed assignment described in \Sref{sec:init} with the following two approaches.

\noindent\textbf{All negative}: In a similar way to the existing PU learning methods, all the unlabeled data are regarded as negative samples. In this case, the initial labels contain noisy labels with a noise rate of $\pi_\mathrm{p}$.

\noindent\textbf{Randomized hard-labels}: Initial pseudo-labels are assigned as follows:
\begin{equation}
    y^{\mathrm{rand}}_i=
    \begin{cases}
    \text{$0$ with the probability of $\pi_\mathrm{n}$}\\
    \text{$1$ with the probability of $\pi_\mathrm{p}$}
    \end{cases}
    .
\end{equation}
The obtained initial labels $\{y_i^{\mathrm{rand}}\}_{i=1}^{n_\mathrm{u}}$ satisfies the class prior probability: $\frac{1}{n_\mathrm{u}}\sum_{i=1}^{n_\mathrm{u}}y_i^{\mathrm{rand}} = \pi_\mathrm{p}$.
In this case, the noise rate is $1-\pi_\mathrm{n}^2-\pi_\mathrm{p}^2$.

We show the performance of the approaches presented in this research using different initial pseudo-labels in \Tref{tab:results0}.
The proposed assignment of initial pseudo-labels always outperforms other methods. These results indicate that beginning by considering all the unlabeled data as negative samples is not effective and that considering the class prior probability is important.

\begin{table}[t]
    \caption{Mean and standard deviation of recovery error rates for unlabeled training data over 10 trials for MNIST, CIFAR-10, and 20 Newsgroups. Different methods are compared using the same 10 random samplings.}
    \begin{center}
    \begin{tabular}{c|c|c|c}\hline
    \multicolumn{1}{c|}{Dataset}&\multicolumn{1}{c|}{nnPU}&\multicolumn{1}{c|}{PUbN\textbackslash N}&\multicolumn{1}{c}{Our method}\\\hline
    \multicolumn{1}{c|}{MNIST}&$7.06\pm1.14$&$5.04\pm0.60$&$\bm{3.88}\pm\bm{0.68}$\\\hline
    \multicolumn{1}{c|}{\begin{tabular}{c}CIFAR-10\vspace{-0.5mm}\\(vehicles as P)\end{tabular}}&$12.15\pm0.52$&$10.37\pm0.36$&$\bm{9.45}\pm\bm{0.34}$\\\hline
    \multicolumn{1}{c|}{\begin{tabular}{c}CIFAR-10\vspace{-0.5mm}\\(mammals as P)\end{tabular}}&$21.88\pm1.06$&$20.85\pm0.69$&$\bm{19.52}\pm\bm{0.77}$\\\hline
    \multicolumn{1}{c|}{20 Newsgroups}&$13.41\pm0.81$&$12.27\pm0.76$&$\bm{11.78}\pm\bm{0.69}$\\\hline
    \end{tabular}
    \end{center}
    \label{tab:resultsr}
\end{table}

\begin{table}[t]
    \caption{Mean and standard deviation of test error rates over 10 trials for MNIST under different class prior probabilities. Different methods are compared using the same 10 random samplings.}
    \begin{center}
    \begin{tabular}{c|c|c|c}\hline
    \multicolumn{1}{c|}{Class prior probability}&\multicolumn{1}{c|}{nnPU}&\multicolumn{1}{c|}{PUbN\textbackslash N}&\multicolumn{1}{c}{Our method}\\\hline
    \multicolumn{1}{c|}{$\pi_\mathrm{p}=0.3$}&$5.11\pm0.86$&$5.15\pm0.79$&$\bm{3.60}\pm\bm{0.39}$\\\hline
    \multicolumn{1}{c|}{$\pi_\mathrm{p}=0.4$}&$5.58\pm0.98$&$4.83\pm0.88$&$\bm{3.47}\pm\bm{0.43}$\\\hline
    \multicolumn{1}{c|}{$\pi_\mathrm{p}=0.5$}&$6.17\pm1.08$&$5.29\pm0.90$&$\bm{4.10}\pm\bm{0.80}$\\\hline
    \multicolumn{1}{c|}{$\pi_\mathrm{p}=0.6$}&$8.01\pm1.52$&$5.54\pm0.87$&$\bm{5.12}\pm\bm{0.88}$\\\hline
    \multicolumn{1}{c|}{$\pi_\mathrm{p}=0.7$}&$11.57\pm2.00$&$7.13\pm1.37$&$\bm{6.16}\pm\bm{0.85}$\\\hline
    \end{tabular}
    \end{center}
    \label{tab:results2}
\end{table}

\subsubsection{Comparison with the state-of-the-art methods}
We have compared the proposed method with two state-of-the-art PU learning methods: non-negative PU learning (nnPU)~\cite{kiryo2017positive} and PUbN\textbackslash N~\cite{hsieh2018classification}.

\Tref{tab:results} shows the test error rates of the comparison; the proposed method outperforms other methods.
\Tref{tab:resultsr} shows the recovery error rates for unlabeled training data of the comparison. This indicates that the proposed method can identify negative (and positive) samples from unlabeled data with high accuracy.


Unlike most existing two-step approaches, including PUbN\textbackslash N, which use only reliable negative samples from unlabeled data, our method can fully utilize unlabeled data because of cleaning of noisy-labeled data.

\begin{figure*}[t]
    \centering
    \subfloat[MNIST]{%
    \includegraphics[width=8.8cm,height=7cm]{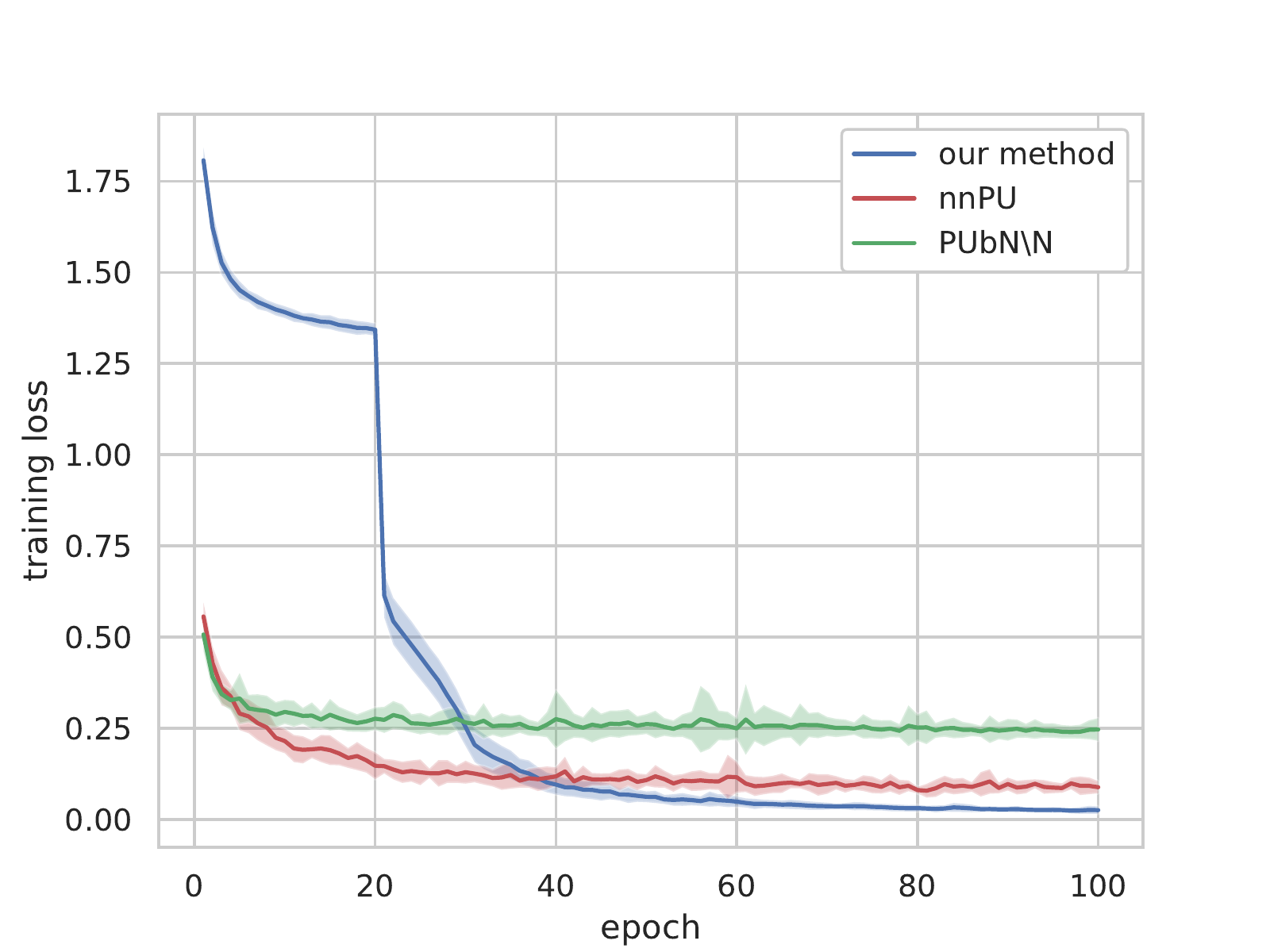}}%
    \subfloat[CIFAR-10 (vehicles as P)]{
    \includegraphics[width=8.8cm,height=7cm]{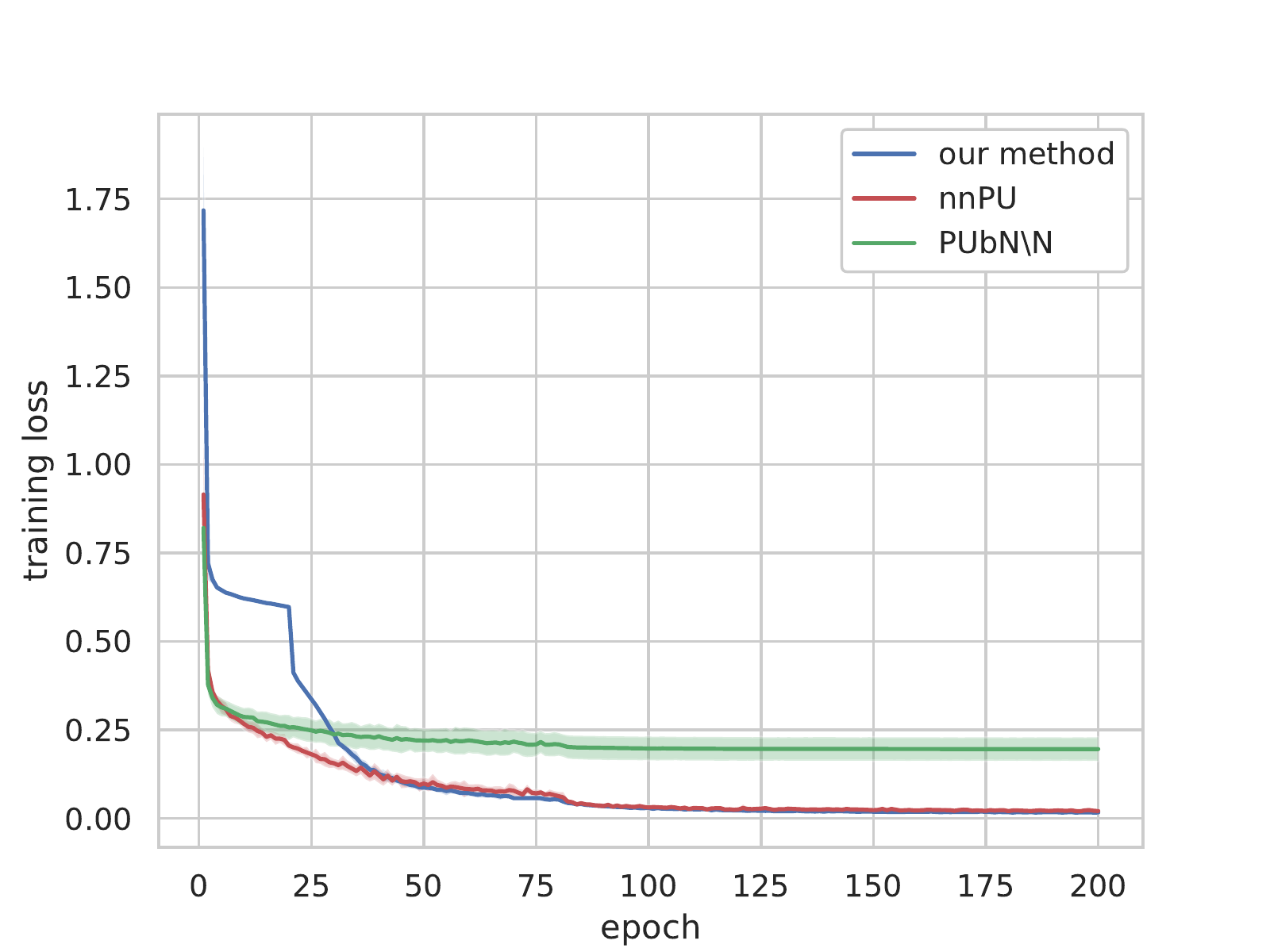}}
    \\
    \subfloat[CIFAR-10 (mammals as P)]{
    \includegraphics[width=8.8cm,height=7cm]{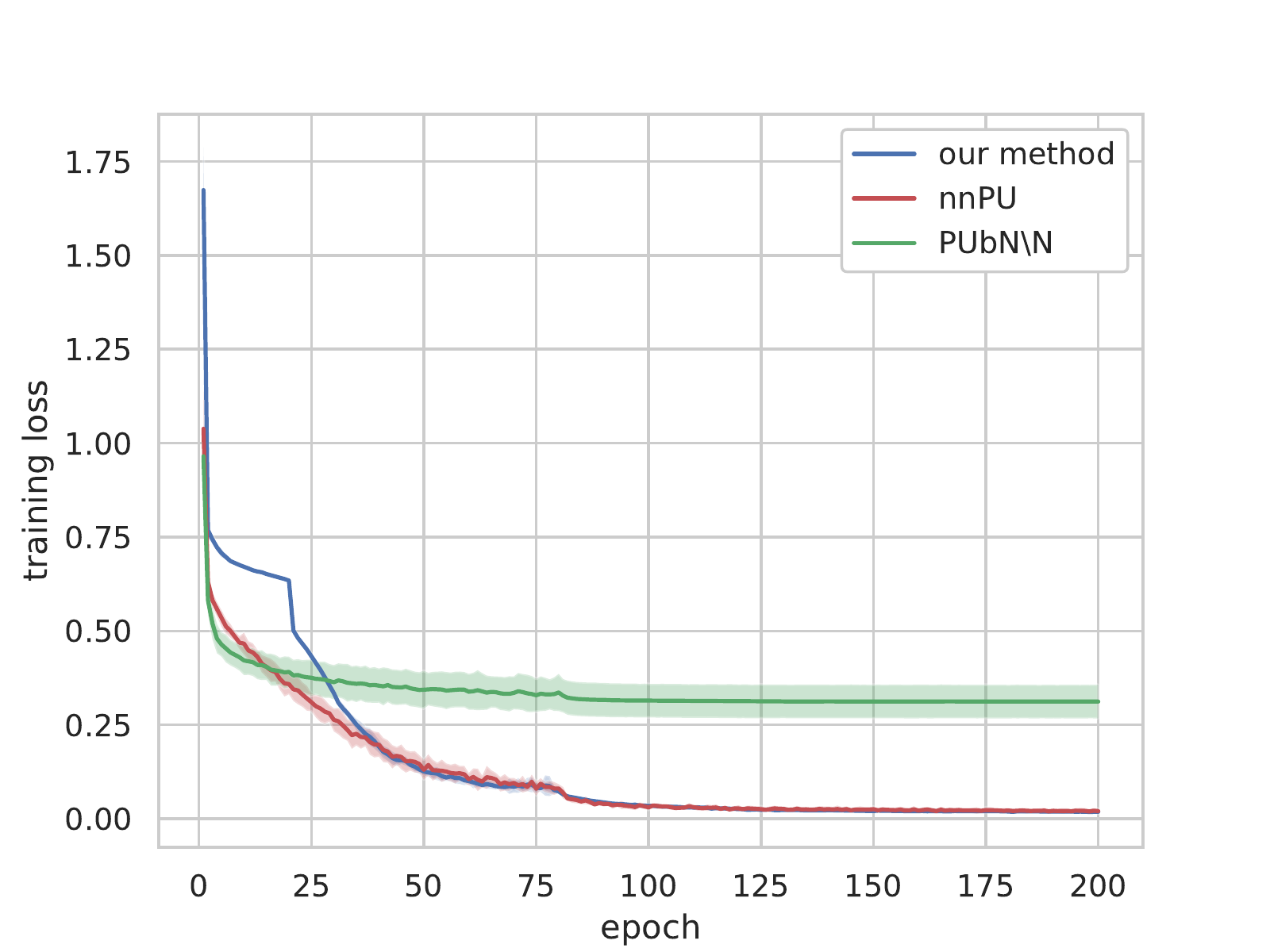}}
    \subfloat[20 Newsgroups]{
    \includegraphics[width=8.8cm,height=7cm]{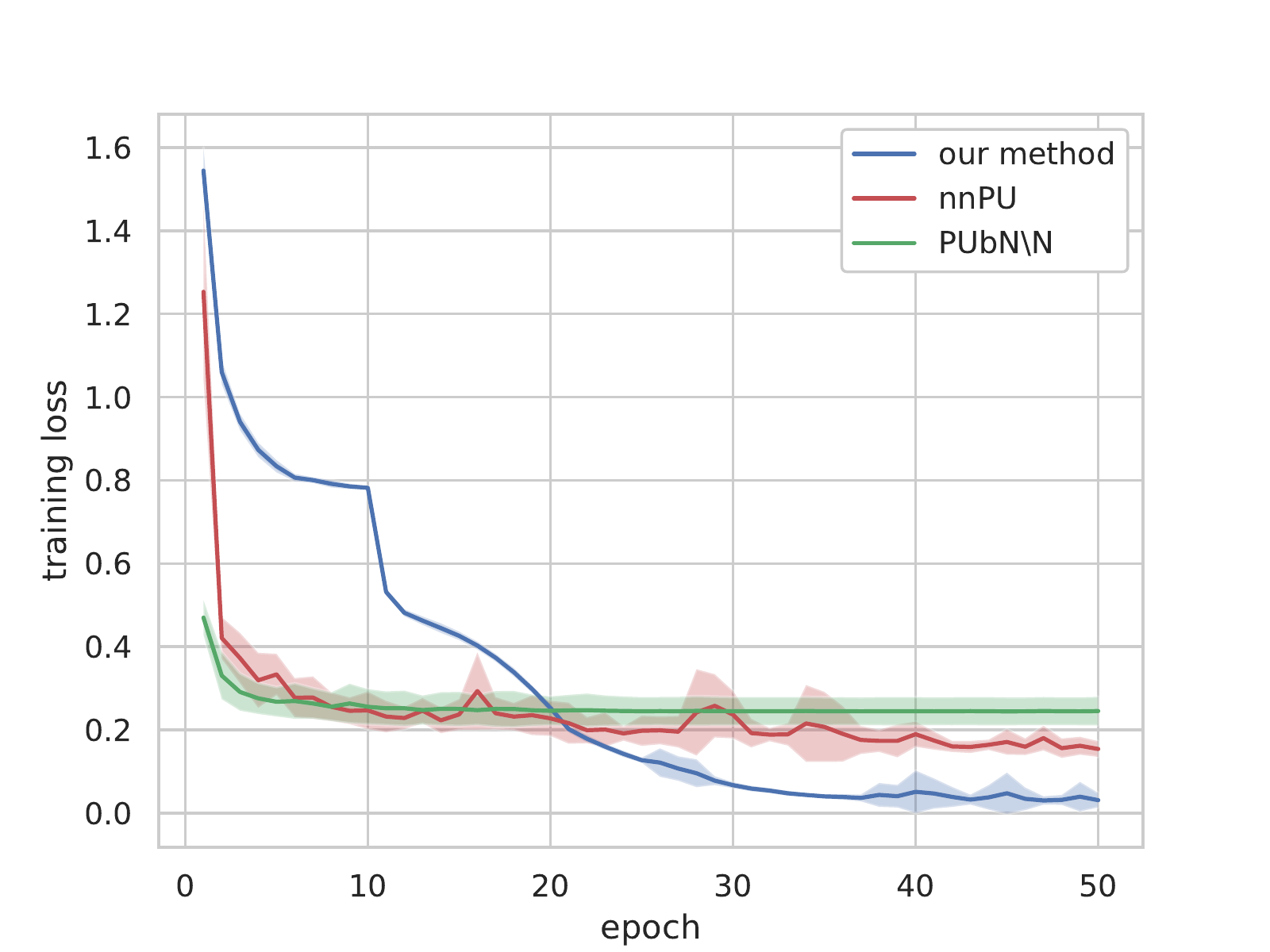}}
    \caption{Training loss curve comparison of means and standard deviations computed by nnPU, PUbN\textbackslash N, and the proposed method (the first step and the second step) under the same 10 random samplings of MNIST, CIFAR-10 (the vehicles are the positive class), CIFAR-10 (the mammals are the positive class), and 20 Newsgroups.}
    \label{fig:graph}
\end{figure*}

We show the training loss curve in \Fref{fig:graph}.
In all the experiments, updating the labels of unlabeled data started at the $e_{\mathrm{start}}=20$th epoch, and it is observed that the training losses did not drop steeply before this epoch.
This means that the classifiers did not overfit to the initial pseudo-labels because we carefully assigned them as described in~\Sref{sec:init}.

\subsubsection{Performance on different class prior probabilities}
We have further investicated certain cases, varying the class prior probability $\pi_\mathrm{p}$, in order to examine how $\pi_\mathrm{p}$ affects the performance. 
We have moderately reduced positive or negative examples to prepare unlabeled datasets with the class prior probability being in $\{0.3, 0.4, 0.5, 0.6, 0.7\}$.
The experimental results are displayed in \Tref{tab:results2}. The results show that the proposed method consistently outperforms the other two methods, leading to the conclusion that the method presented in this research is more robust for the imbalance of the class.

\section{Conclusion}
In this study, as compared to existing PU learning, we have introduced a different perspective regarding the treatment of unlabeled data. We have considered unlabeled data as noisy-labeled data, and introduced a new approach to PU learning in which the network and noisy labels are jointly optimized.

Unlike noisy-labeled learning, clean-labeled data has only positive samples in PU learning and thus, we have proposed a new weighting parameter to emphasize positive samples in the beginning of training.
Then, we have determine the best initial label assignment by considering the class prior probability.
Experimental results demonstrate that the proposed method significantly outperforms the state-of-the-art methods on MNIST, CIFAR-10, and 20 Newsgroups datasets.


%




\ifCLASSOPTIONcaptionsoff
  \newpage
\fi



\bibliographystyle{IEEEtran}
\bibliography{references}
\end{document}